\documentclass{article}

\PassOptionsToPackage{numbers, compress}{natbib}
\usepackage[preprint]{neurips_2020}

\usepackage[british]{babel}     % follow appropriate hyphenation rules
\usepackage[utf8]{inputenc}     % allow utf-8 input
\usepackage[T1]{fontenc}        % use 8-bit T1 fonts
\usepackage{url}                % simple URL typesetting
\usepackage{booktabs}           % professional-quality tables
\usepackage{amsfonts}           % blackboard math symbols
\usepackage{nicefrac}           % compact symbols for 1/2, etc.
\usepackage{microtype}          % microtypography

\usepackage{amsmath}
\usepackage{amssymb}
\usepackage{xcolor}
\usepackage{colortbl}
\usepackage{mathtools}
\usepackage{isomath}
\usepackage{multirow}
\usepackage{algorithm, algpseudocode}
\usepackage{caption}
\usepackage{subcaption}
\usepackage{wrapfig}
\usepackage{graphbox}           % Adds align option to graphicx
\usepackage{soul}               % Adds commands aimed at emphasising such as
\usepackage{selectp}
\usepackage{hyperref}           % (usually) has to be the last package to be imported

\DeclareMathOperator*{\argmin}{arg\,min}

\DeclarePairedDelimiter\abs{\lvert}{\rvert}%

\DeclareMathOperator{\mask}{\tensorsym{M}}
\DeclareMathOperator{\smask}{\tilde{\tensorsym{M}}}

\definecolor{ourmethod}{gray}{0.93}

\title{Single Shot Structured Pruning Before Training}

\author{%
  Joost van Amersfoort\thanks{Equal contribution} \And
  Milad Alizadeh\footnotemark[1] \And
  Sebastian Farquhar\footnotemark[1] \AND \vspace{5mm}
  Nicholas Lane \hspace{5mm} Yarin Gal \\
  Department of Computer Science \\
  University of Oxford \\
  \texttt{\{joost.van.amersfoort, milad.alizadeh, sebastian.farquhar\}@cs.ox.ac.uk}
}

\begin{document}
\maketitle

%!TEX root = ../neurips_2020.tex
\begin{abstract}
    We introduce a method to speed up training by 2x and inference by 3x in deep neural networks using structured pruning applied \emph{before training}.
    Unlike previous works on pruning before training which prune individual weights, our work develops a methodology to remove entire channels and hidden units with the explicit aim of speeding up training and inference.
    We introduce a compute-aware scoring mechanism which enables pruning in units of sensitivity per FLOP removed, allowing even greater speed ups.
    Our method is fast, easy to implement, and needs just one forward/backward pass on a single batch of data to complete pruning before training begins.
\end{abstract}

%!TEX root = ../neurips_2020.tex

\section{Introduction}\label{sec:introduction}

Pruning techniques \citep{obd,obs,eigendamage} are able to successfully compress and speed up inference in trained neural networks.
However they do nothing to address the speed and computational cost of \emph{training} the initial model, which can have a significant CO$_2$ footprint \citep{strubell_energy_2019}.
Recently, it has been shown that it is possible to prune a deep network \emph{before training} \citep{lee2018snip, wang_picking_2019}.
These methods are \emph{unstructured} pruning methods, i.e., they prune individual weights from convolutional or linear layers.
Unstructured pruning methods only lead to speed improvements with specialized hardware \citep{sze2017efficient}, and because sparse weights do not induce sparse activations, they do not reduce run-time memory footprint either.

In this work, we propose a \emph{structured} and \emph{compute-aware} Pruning Before Training (PBT) method.
Structuring pruning methods remove entire channels in convolutional layers and hidden units in linear layers, leading to speed ups and reduced memory consumption on standard compute devices.
Our method, Single Shot Structured Pruning (3SP), is easy to implement and has few hyper-parameters to tune.
The pruned model trains 2x faster (on a GPU) and performs inference 3x faster (on a CPU), with only a 0.5\% loss in accuracy on CIFAR-10.
Our method needs just one forward-backward pass through the model with a single minibatch of data.
To the best of our knowledge, structured pruning has never been attempted before training.

3SP is based on the SNIP objective \citep{lee2018snip}, and we give an extensive empirical analysis of extending this objective to the structured setting.
We further introduce a compute-aware weighting of the pruning score which measures the impact on the loss \emph{per unit compute removed}.
This actively biases pruning to remove more compute-intensive channels: a channel could be removed if either it has a small effect on the loss, or if it has a significant computational cost.
Using this additional term, we are able to increase compute reduction from 60\% to 85\%.

Speeding up the training of large neural networks is useful when the training data changes quickly and models need to adjust rapidly.
One example of this is active learning \citep{Settles2010}, where trained models are used to identify the most informative datapoints to label for inclusion in the training data.
In \S\ref{sec:active_learning} we show how 3SP can be used to identify valuable data to acquire faster than a full model, allowing us to achieve \emph{better} accuracy within a time-budget than an un-pruned model could.

Our main contributions are:
\begin{itemize}
    \item We introduce 3SP, a structured pruning before training (PBT) method that speeds up training by 2x and inference by 3x with only a small loss in accuracy.
    \item We show how to prune explicitly in units of compute cost, rather than number of weights, allowing even greater reduction in compute.
    \item We study empirically different objectives for unstructured pruning before training (SNIP and GraSP) and study the impact of moving to structured pruning domain.
\end{itemize}

%!TEX root = ../neurips_2020.tex
\begingroup % So that I can locally set \intextsep
\setlength{\intextsep}{0pt}
\section{Background}\label{sec:background}
\begin{wrapfigure}[16]{R}{0.49\textwidth}
   \vspace{-6mm}
   \captionsetup{labelfont={bf}}
   \centering
   \includegraphics[width=0.49\textwidth]{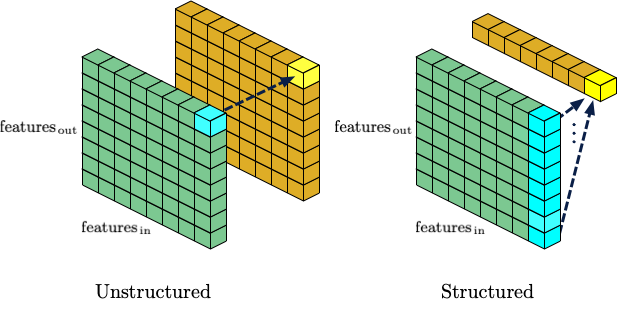}
   \caption{
      Binary mask tensor (orange) in structured vs. unstructured pruning for a fully-connected layer's weight matrix (green).
      There is one mask per weight in the unstructured pruning mask $\tensorsym{M}$, while in structured pruning, the binary mask $\smask$ has one entry per hidden unit (a column of the weight tensor).}
   \label{fig:struc-vs-unstruc-pruning}
\end{wrapfigure}
In this section, we review key concepts that 3SP builds upon, leaving a comparison to other model compression techniques in the literature to \S\ref{sec:related_work}.

\textbf{Pruning Before Training.}
SNIP \citep{lee2018snip} introduced an effective method to prune individual weights before training.
SNIP aims to remove weights from the neural network such that the difference between the loss of the full model and the loss of the pruned model is as small as possible.
This is the same goal as the classic post-training pruning approaches in Optimal Brain Damage \citep{obd} and Optimal Brain Surgeon \citep{obs}, but applied \emph{before training}.
To approximate the effect on the loss of removing a weight, SNIP attaches a multiplicative binary mask to each weight.
The mask value is one when a weight is kept as part of the model and zero when it is pruned.
Their method can therefore be thought of as finding the value of the binary mask tensor $\mask$ which minimizes the change in loss:

\begin{equation}
   \argmin_{\mask} \abs*{\mathcal{L}\big(f_{\mathbf{w} \odot \mask}(\mathbf{x}), y\big) - \mathcal{L}\big(f_{\mathbf{w}}(\mathbf{x}), y\big)}, \label{eq:snip}
\end{equation}

with $f_{\mathbf{w} \odot \mask}(\cdot)$ the neural network evaluated with weights $w$ masked by mask $\mask \in \{0,1\}^{|w|}$.
This discrete optimization problem is intractable.
Instead \citet{lee2018snip} solve a continuous relaxation by differentiating the loss with respect to the mask parameters on a single batch of training data and using the first-order approximation: $\mathcal{S}_\text{SNIP} = \frac{\partial \mathcal{L}}{\partial \mask}\big\rvert_{\mask=\mathbf{1}}$.
Using the tensor of scores, $\mathcal{S}_\text{SNIP}$, a threshold is computed given the desired prune ratio.
All entries below the threshold are set to zero.
SNIP relies on the strong assumption that the importance of the weights does not depend on which other weights are removed, i.e. no higher-order terms in the Taylor expansion are considered.
\endgroup

GraSP \citep{wang_picking_2019}, an alternative PBT method, uses a different criterion based on preserving gradient flow.
GraSP keeps weights that contribute to the norm of the gradient, removing weights can potentially even improve gradient flow.
While it works well in unstructured pruning, especially at high levels of sparsity, we show in \S\ref{sec:grasp} that GraSP's approximation has shortcomings which are particularly problematic for structured pruning.
In the rest of this paper, we therefore focus on the SNIP criterion.

\textbf{Structured Pruning.}
In structured pruning, only entire channels in convolutional layers and columns of linear layers can be removed (Figure \ref{fig:struc-vs-unstruc-pruning}).
This is a significant restriction compared to unstructured pruning where any individual weight can be removed.
However, structured pruning reduces the computational cost of training and evaluating a model, because when entire channels are removed the size of the activations is also reduced, leading to a smaller model.
Removing single weights does not lead to reduced (or even sparse) activations, and speeding up the computation of sparse convolutions and linear layers requires specialized hardware \citep{sze2017efficient}.
In contrast, computational benefits from structured pruning are straightforward: the architecture is smaller.

The assumption that the importance of weights are independent of each other, as described in the previous section, is even stronger in the structured case.
Removing a single (output) channel can, for example, lead to the removal of 512 x 3 x 3 weights, such as in the later layers of VGG \citep{simonyan2014very} and ResNet \citep{he_deep_2016}.
Extending any unstructured method to a structured method is therefore not trivial, especially when the original score is noisy as in pruning before training.

%!TEX root = ../neurips_2020.tex

\begin{figure}[t]
    \centering
    \begin{subfigure}[b]{0.49\textwidth}
        \centering
        \includegraphics[width=\textwidth]{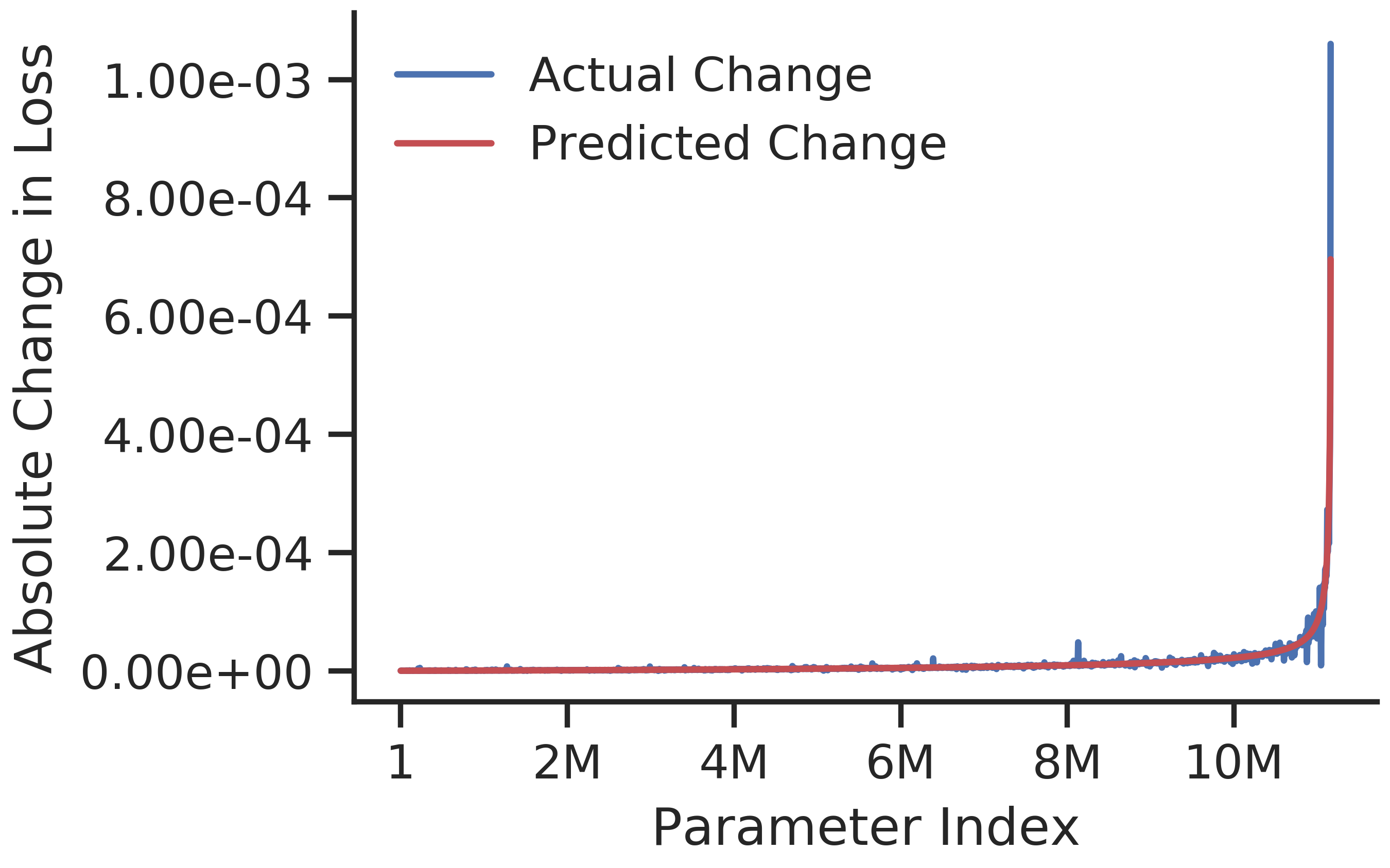}
        \caption{Unstructured}
    \end{subfigure}
    \hfill
    \begin{subfigure}[b]{0.49\textwidth}
        \centering
        \includegraphics[width=\textwidth]{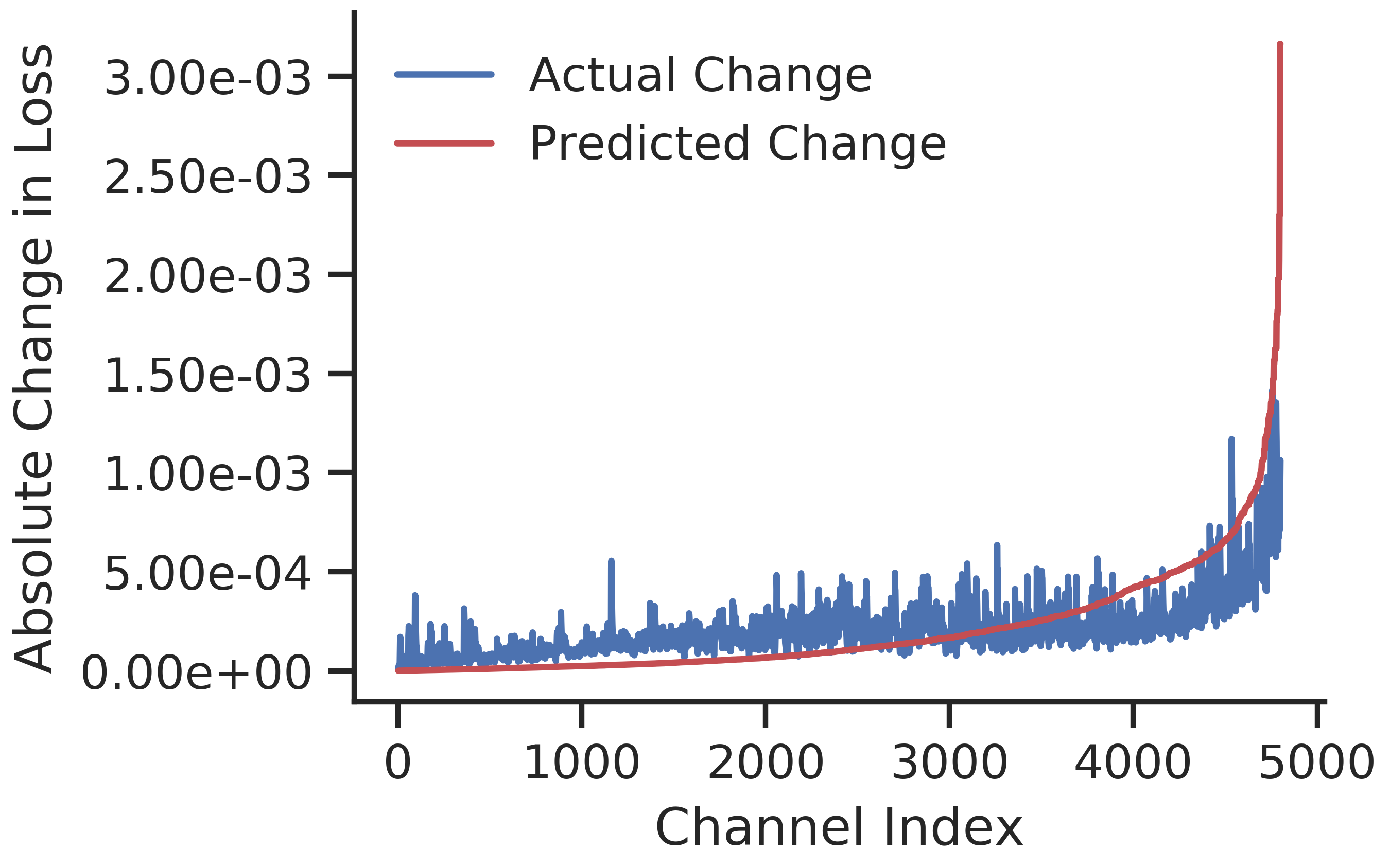}
        \caption{Structured}
    \end{subfigure}
    \caption{
        Approximation accuracy of scores in unstructured and structured pruning.
        We compare the predicted change in loss from the Taylor expansion (scores) with the actual change in loss (normalized).
        The X-axis is sorted by the scores.
        While scores in structured pruning are noisy, we can still successfully identify parameters that contribute the most to the change in the loss.
    }
    \label{fig:pred_vs_actual_snip}
\end{figure}
\section{Method}\label{sec:method}

Our method 3SP introduces a structing PBT method, based on the unstructured SNIP objective, enabling a significant improvement in training and inference speed using only a single forward-backward pass on one minibatch of data.
We further provide a compute-aware extension to 3SP which aims to optimize the mask tensor in units of compute cost rather than numbers of weights.
We provide a step by step description of how 3SP and its compute-aware extension works in Algorithm \ref{algorithm}.
\begin{algorithm}[t]
    \begin{algorithmic}[1]
        \Statex \textbf{Data:} minibatch of data from dataset $\mathcal{D}$.
        \Statex \textbf{Result:} Best model mask $\smask^*$ for pruned model.
        \Statex \textbf{Initialization:} NN with in channels $C^\text{in}_l$, spatial height and width $H_l$ and $W_l$ and kernel size $K_l$ for each layer $l$, initialized with \citet{he_deep_2016}; $\smask$ initialized at 1 for every entry; target pruning ratio $p$; compute smoothing term $\lambda$; $\tensorsym{R} \in \mathcal{R}^{|\smask|}$;
        \State $g \leftarrow \frac{\partial\mathcal{L}}{\partial\smask}$ \Comment{$1^{st}$ order apprx.\ $\mathcal{S}$ using one minibatch.}
        \If{Compute-Aware}
        \State $c_l \leftarrow C_{\text{in}} \cdot H_l \cdot W_l \cdot K_{l}^2$ \Comment{Calculate cost-per-layer.}
        \State $\bar{c}_l \leftarrow \frac{c_l + \lambda}{\sum_j (c_j + \lambda)}$ \Comment{Apply compute cost smoothing.}
        \State $\bar{c}_l \leftarrow \frac{\bar{c}_l}{\max_j (\bar{c}_j)}$
        \Comment{Normalize compute cost.}
        \For{Score $g_i$ and associated layer $l$}
        \State$\tensorsym{R}_i \leftarrow \frac{\abs{g_i}}{\bar{c}_l}$ \Comment{Convert 3SP score to cost-space}
        \EndFor
        \Else
        \State$\tensorsym{R} \leftarrow \abs{g_i}$
        \Comment{Non-compute-aware 3SP score.}
        \EndIf
        \State $\tensorsym{R}_{\text{threshold}} \leftarrow$ $p$'th percentile of entries in $\tensorsym{R}$\;
        \State $\smask^* \leftarrow \tensorsym{R} \geq \tensorsym{R}_{\text{threshold}}$ \Comment{Keep channels/units with high score.}
    \end{algorithmic}
    \caption{Structured Compute-Aware Pruning - 3SP}
    \label{algorithm}
\end{algorithm}
\subsection{Structured Pruning Before Training}
SNIP defines $\mask$ with the same shape as the weights, allowing it to turn off individual weights.
We instead define $\smask$ to remove entire operations.
In particular, for convolutional layers, each output channel gets one binary mask variable governing its entire spatial extent.
Linear layers have a binary mask per hidden unit; we visualize this in Figure \ref{fig:struc-vs-unstruc-pruning}.
Masking, therefore, can be implemented by changing the shape of the weight tensors and is equivalent to using a \textit{smaller model}, unlike unstructured methods.
We minimize a similar objective to SNIP:
\begin{equation}
    \argmin_{\smask} \abs*{\mathcal{L}\big(f_{\smask}(\mathbf{x}), y\big) - \mathcal{L}\big(f_{\mathbf{w}}(\mathbf{x}), y\big)}, \label{eq:s-snip}
\end{equation}
where $f_{\smask}$ is the model given by masking channels using $\smask$, in practice this is done by removing all the weights associated with masked operations from the model leading to a smaller, faster model.
This is in contrast to unstructured pruning where the mask is applied using an element-wise product.

\textbf{3SP assumptions.} We approximate this change in loss with three assumptions: \textbf{1.} We approximate the binary mask $\smask$ as a continuous variable; \textbf{2.} We use a first-order Taylor expansion---$\mathcal{S} = \frac{\partial\mathcal{L}}{\partial\smask}\big\rvert_{\smask=\mathbf{1}}$---approximated with just one minibatch of data; \textbf{3.} (Given 2.) We approximate the change implied by changing a set of entries of $\smask$ as being the sum of individual changes.

These assumptions are similar to the unstructured SNIP assumptions but note that in the structured setting, these approximations are stronger, requiring additional justifications.
Unstructured models have a mask variable with potentially millions of entries (the number of weights), while a structurally pruned VGG-19 has an $\smask$ with roughly 5,000 entries.
This means that the interaction between any two entries will often be much larger, making assumptions 2 and 3 less plausible.
In Figure \ref{fig:pred_vs_actual_snip}, we assess how appropriate the first-order Taylor approximation is for predicting the change in loss when removing individual weights.
We show on the left that the unstructured SNIP approximation is close to the actual change in the loss when a single weight is removed.
(For the unstructured curve, we compute the actual change in loss for every five thousandth weight, rather than computing all of the millions of scores.)
On the right, we show that in the structured case, there is significantly more noise, but the predicted change in loss correlates strongly with the actual change.
We discuss this effect and a comparison to GraSP in more detail in \S\ref{sec:grasp}.

\textbf{Rescaling Initializations.} Pruning before training changes the variance of activations at initialization and in the structured case even changes the number of activations.
Previous research into initialization schemes for neural networks \citep{he2015delving} shows the importance of the variance and the number of activations for model training.
By pruning a model, we reduce the number of output channels of most layers, which means the variance no longer has the right value relative to the number of output activations.
We attempt to correct for this by studying the effects of rescaling weights by the ratio: $\sqrt{\frac{\text{original fan out}}{\text{pruned fan out}}}$
which amounts to recovering the variance scaling suggested by \citet{he2015delving}.
However, we note that because large and small weights are not pruned uniformly, the resulting variance of the model weights of the pruned model is not exactly what one would get from initializing the model from scratch.
The authors of SNIP reinitialize their models after pruning, which may address the same issue.

\subsection{Compute-Aware 3SP (3SP + CA)}
Different layers in a model have a different computationtal cost. For example the computational cost in FLOPs of a convolution operation is $2 \cdot H \cdot W \cdot C^\text{ out} \cdot C^\text{ in} \cdot K^{2}$ where $H$ and $W$, are the height and width of the output, $C_{\text{out}}$, $C_{\text{in}}$ are the number of out and in channels, and $K$ is the size of the kernel.
In earlier layers of the model, the spatial dimensions tend to be high, while in later layers the number of input channels increase.
Since our aim is to remove as much compute from the model while preserving model accuracy, we extend 3SP to be compute-aware, by dividing the score by the normalized compute cost $c_i$ per channel:
\begin{equation}
    \bar{c}_i = \frac{c_i}{\sum_j c_j} \qquad \text{where} \qquad c_i \coloneqq 2 \cdot H_i \cdot W_i \cdot C_i^\text{ in} \cdot K_{i}^2.
\end{equation}
We calculate a retention score, $r_{i,j}$, for layer $i$, channel $j$, corresponding to each mask entry, which measures the impact on the loss per unit of compute removed:
\begin{equation}
    r_i = \frac{\mathcal{S}_{i,j}}{\tilde{c}_i} \label{eq:retain}.
\end{equation}
This retention score can be high either if a channel has a big effect on the loss, or if the channel has a negligible effect on the compute cost.
A low retention score means that the channel is either harmful to prune or would offer little speed-up.
In practice, we would like to trade off the importance of compute and change in the loss, so we introduce a Laplace smoothing parameter $\lambda$ \citep{manning_introduction_2009}: $\tilde{c}_i = \frac{\bar{c}_i + \lambda}{\sum_j \bar{c}_j + \lambda}.$
A larger values of $\lambda$ makes the pruning depend more on the predicted change in loss and pay less attention to the compute costs of different layers.
%!TEX root = ../neurips_2020.tex
%
\begin{figure}[t]
    \begin{minipage}[t]{0.49\textwidth}
        \includegraphics[width=0.95\linewidth]{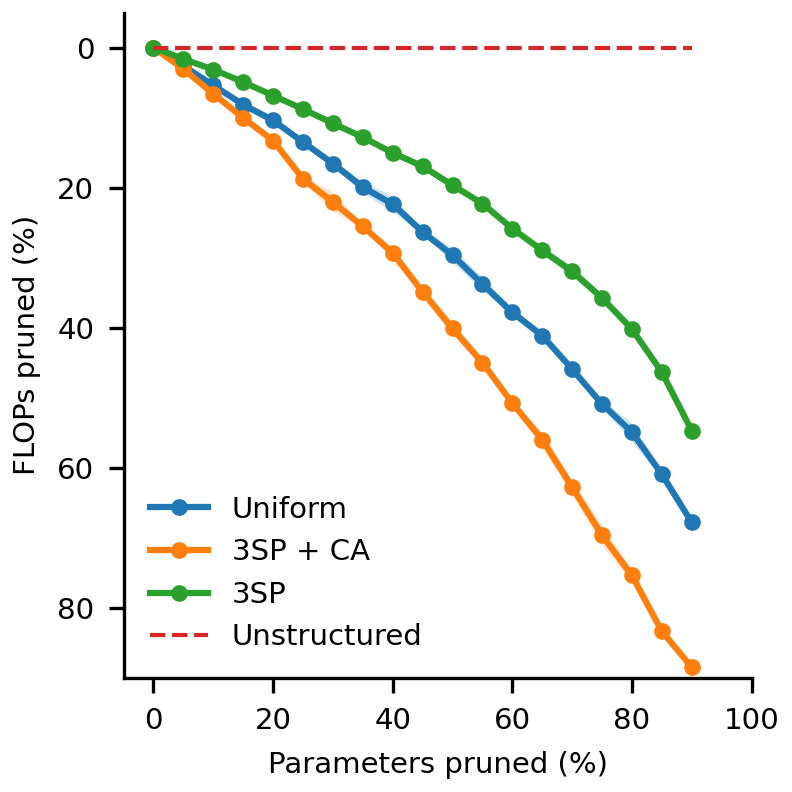}
        \caption{
            The relation between pruning parameters and the effect on compute on CIFAR-10 with VGG-19.
            Uniform is a naive baseline which prunes all channels with uniform probability, and therefore ignores both model loss and compute cost. 3SP removes channels least important to the loss, which happen to also use less compute on average.
            3SP + CA, without compute smoothing, actively prunes FLOPs while taking model performance into account (see Figure \ref{fig:tradeoff_accuracy_mac}).
        }
        \label{fig:compute_improvement}
    \end{minipage}
    \hfill
    \begin{minipage}[t]{0.49\textwidth}
        \includegraphics[width=0.95\linewidth]{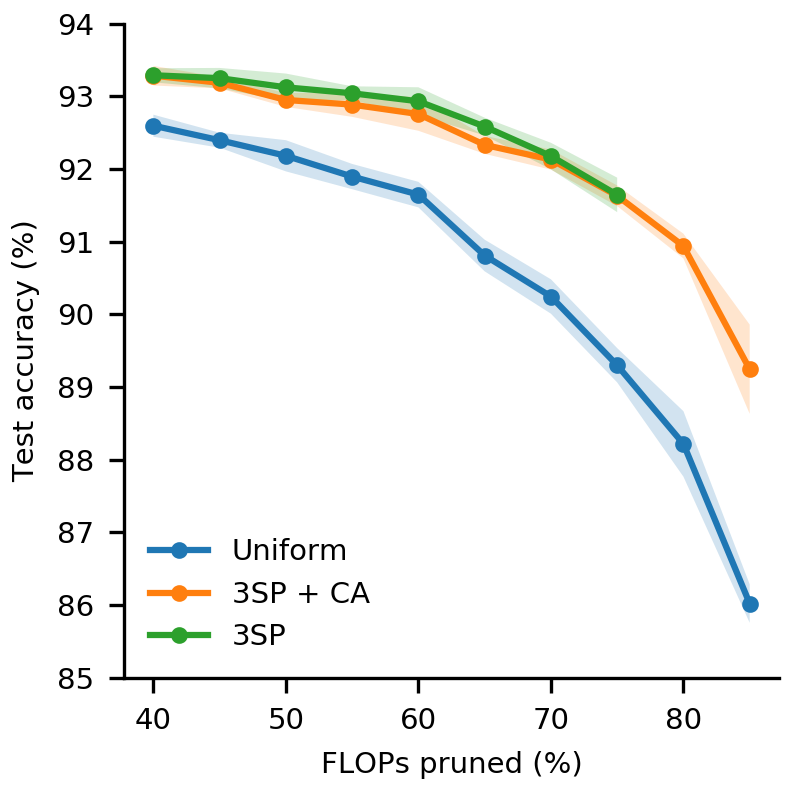}
        \caption{
            The performance of 3SP and 3SP + CA (compute-aware) against a baseline of uniformly removing channels from the model on CIFAR-10 with VGG-19.
            At every amount of compute pruned, 3SP outperforms uniform, with the difference increasing as more FLOPs are pruned.
            After removing 75\% of FLOPs, 3SP becomes unable to prune more without pruning entire layers.
            Our compute-aware extension, however, is able to remove even more compute without failing.
        }
        \label{fig:tradeoff_accuracy_mac}
    \end{minipage}
\end{figure}
\section{Experiments}\label{sec:experiments}
In this section, we show that models pruned with 3SP are substantially faster than unpruned models and models pruned using baseline PBT methods without sacrificing much in accuracy.
We show that a compute aware version (3SP + CA) can be even faster, but with a more significant reduction in performance.
Finally, we demonstrate how this capability might be beneficial in a pratical setting where training speed is a bottleneck: we show that a 3SP model can achieve higher accuracy than a full model with a time-budget in active learning.

\textbf{Baselines.} We compare to previously published unstructured PBT methods SNIP \citep{lee2018snip} and GraSP \citep{wang_picking_2019}, though we note that these methods do not provide a speed-up.
We do not compare to methods that prune after or during training, because these incur a significant upfront computational cost, while this paper is focussed on constraining the \emph{training} cost.
3SP and its compute-aware extension are the first methods to structurally prune a model before training, showing that this approach is feasible and a ground for further research.

We also compare to a naive structured pruning method which uniformly prunes over all mask entries and layers.
This ``Uniform'' baseline samples entries for binary mask tensors $\mask$ or $\smask$ from $\text{Bern}(p)$ in order to prune a $p$'th of the weights.
In expectation, the model width is pruned by a uniform ratio in each layer.
This is similar to using a narrower model.
\citet{rethinking_the_value_of_pruning} showed that these sorts of untargeted pruning methods are effective baselines that are able to obtain strong performance.

\textbf{Architecture and random seeds.} Our experiments are executed using VGG-19; we describe the exact architecture in Appendix~\ref{app:architecture}.
We run all our experiments 5 times using different random seeds, reporting the mean and standard error of each experiment in tables (standard deviation in figures, as s.e. was not visible).

\textbf{Measuring compute.} We report model compute cost in Floating Point operations (FLOPs).
The exact number of FLOPs used to perform a calculation depends on the hardware, so we make the common assumption that roughly two FLOPs are required for each multiply-accumulate operation.

\textbf{Compute-cost vs. Accuracy Trade-off.}
In Figure \ref{fig:compute_improvement} we show that our approach is able to reduce the FLOPs required for a forward pass in the model.
3SP tries to preserve model performance, so it removes compute less aggressively than the Uniform baseline.
However, 3SP + CA (compute-aware) with $\lambda=0$ is able to very aggressively remove compute.
In Figure \ref{fig:tradeoff_accuracy_mac} we show the trade-off between the compute cost of the model and accuracy for 3SP, 3SP + CA and Uniform pruning on CIFAR10.
We show that there is a significant gap between 3SP and uniform, indicating that we are able to successfully prune our model.
When trying to prune very large amounts of compute, having ignored compute costs during pruning, 3SP is forced to remove entire layers.
Our compute aware extension, however, is able to continue pruning to very high levels of compute sparsity.
\begin{table}[t]
    \centering
    \caption{
        Accuracy of 3SP (grey), SNIP and GraSP for VGG19 on CIFAR-10 and CIFAR-100.
        We evaluate on the basis of pruned parameters because prior PBT methods are unstructured and do not reduce compute cost.
        However, the speed-based evaluation of Figure \ref{fig:tradeoff_accuracy_mac} is a much better view of our method's performance.
        Even on the basis of parameter sparsity, 3SP performs comparably with unstructured methods on CIFAR-10 though a gap forms on the more complex CIFAR-100.
        Rescaling/reinitializing has little effect on the smaller dataset, but appears to matter for CIFAR-100.
        The original (unpruned) model obtains 93.6\% accuracy on CIFAR-10 and 72.5\% on CIFAR-100.
    }
    \resizebox{\textwidth}{!}{
    \begin{tabular}{llccccccccc}
        \toprule
        \multicolumn{2}{c}{\multirow{3}{*}{Method}}              & \multicolumn{3}{c}{CIFAR-10} & \multicolumn{3}{c}{CIFAR-100} \\
        \cmidrule(l){3 - 5}
        \cmidrule(l){6 - 8}
        \multicolumn{2}{c}{} & \multicolumn{3}{c}{Parameters Pruned (Acc. $\pm$ s.e.)} & \multicolumn{3}{c}{Parameters Pruned (Acc. $\pm$ s.e.)} \\
        \cmidrule(l){3 - 5}
        \cmidrule(l){6 - 8}
        \multicolumn{2}{c}{}                                     & 80\%                 & 90\%                  & 95\% & 80\%                 & 90\%                  & 95\%\\
        \midrule
        Unstructured            & Uniform                         & 92.6 $\pm$ 0.04 \%   & 91.4 $\pm$ 0.04 \%    & 89.8 $\pm$ 0.04 \% & 70.3 $\pm$ 0.16 \%  & 68.1 $\pm$ 0.11 \%     & 64.7 $\pm$ 0.27 \%\\
                                & SNIP                           & 93.6 $\pm$ 0.12 \%   & 93.6 $\pm$ 0.05 \%    & 93.4 $\pm$ 0.04 \% & 72.8 $\pm$ 0.10 \%  & 72.4 $\pm$ 0.08 \%     & 70.7 $\pm$ 0.11 \%\\
                                & GraSP                          & 93.2 $\pm$ 0.09 \%   & 93.0 $\pm$ 0.03 \%    & 92.8 $\pm$ 0.08 \% & 71.2 $\pm$ 0.08 \%  & 70.6 $\pm$ 0.15 \%     & 69.5 $\pm$ 0.07 \%\\
        \midrule
        Structured              & Uniform                        & 92.0 $\pm$ 0.08 \%   & 90.4 $\pm$ 0.12 \%    & 89.0 $\pm$ 0.15 \% & 67.5 $\pm$ 0.16 \%  & 63.8 $\pm$ 0.13 \%     & 60.1 $\pm$ 0.29 \%\\
        \rowcolor{ourmethod}
                                & 3SP                            & 93.4 $\pm$ 0.03 \%   & 93.1 $\pm$ 0.04 \%    & 92.5 $\pm$ 0.12 \% & 69.9 $\pm$ 0.14 \%  & 68.3 $\pm$ 0.12 \%     & 63.2 $\pm$ 0.52 \%\\
        \rowcolor{ourmethod}
                                & 3SP + re-init                  & 93.4 $\pm$ 0.04 \%   & 93.0 $\pm$ 0.02 \%    & 92.6 $\pm$ 0.09 \% & 70.3 $\pm$ 0.16 \%  & 69.0 $\pm$ 0.08 \%     & 64.2 $\pm$ 0.35 \%\\
        \rowcolor{ourmethod}
                                & 3SP + re-scale                 & 93.3 $\pm$ 0.03 \%   & 93.0 $\pm$ 0.06 \%    & 92.5 $\pm$ 0.06 \% & 70.5 $\pm$ 0.13 \%  & 69.2 $\pm$ 0.11 \%     & 63.5 $\pm$ 0.63 \%\\
        \bottomrule
      \end{tabular}
    }
      \label{table:CIFAR10}
      \vspace{0mm}
    \end{table}

\textbf{Parameter vs. Accuracy Trade-off.} In Table \ref{table:CIFAR10} we consider accuracy when pruning different proportions of \emph{parameters} using VGG-19 \citep{simonyan2014very} on CIFAR-10 and CIFAR-100.
As we have discussed, removing parameters is not as important as reducing compute cost, which is why the best evaluation of our method is that provided in Figure \ref{fig:tradeoff_accuracy_mac}.
But the only prior PBT methods are unstructured and lead to no compute improvement at all.
Therefore, in order to compare to prior methods, we show in Table \ref{table:CIFAR10} that even when looking at parameter-sparsity, 3SP peforms nearly as well as unstructured methods for CIFAR-10.
At 80\% parameter sparsity 3SP has similar accuracy, and even at 95\% accuracy 3SP is less than a percentage point less accurate than the unstructured SNIP and GraSP methods.
For CIFAR-100, the structured constraint has a bigger effect on performance, though 3SP still performs significantly better than Uniform pruning.
This can be explained by the fact that CIFAR-100 is a more difficult dataset, and VGG19 is therefore less overparameterized.
Re-initializing the weights has a very small effect on the accuracy on CIFAR-10 and helps with CIFAR-100.
This suggests that these pruning before training methods predominantly identify an important model architecture rather than individual weights.

\textbf{Wall Clock Time and Model Size.}
In Table \ref{table:compute_comparison}, we show the tangible benefits of using structured pruning.
The 50\% reduction model is without compute aware, while the 80\% model is with compute aware and smoothing set to $0.05$.
On a GPU, reducing the FLOPs by 80\% or 50\% leads to a 4x or 2x speed-up in training time per epoch.
On a CPU the benefit is even greater.
Because structured pruning results in a much smaller model, and induces a smaller activation map, the forward pass can be kept almost entirely in CPU cache, leading to a 5x or 3x speedup with the same levels of pruning instead.
The smaller memory footprint of our method would allow pruned models to be used in settings that unstructured pruning might not help with.
Adopting our method can therefore help researchers and organizations reduce run-time and power consumption of training, evaluating, and deploying their models, leading to a reduced carbon footprint.
Full details are provided in Appendix \ref{app:wall_clock}.

\begin{wrapfigure}{r}{0.48\textwidth}
    \centering
    \vspace{-12mm}
    \captionsetup{labelfont={bf}}
    \includegraphics[width=2.64in,height=2.3in]{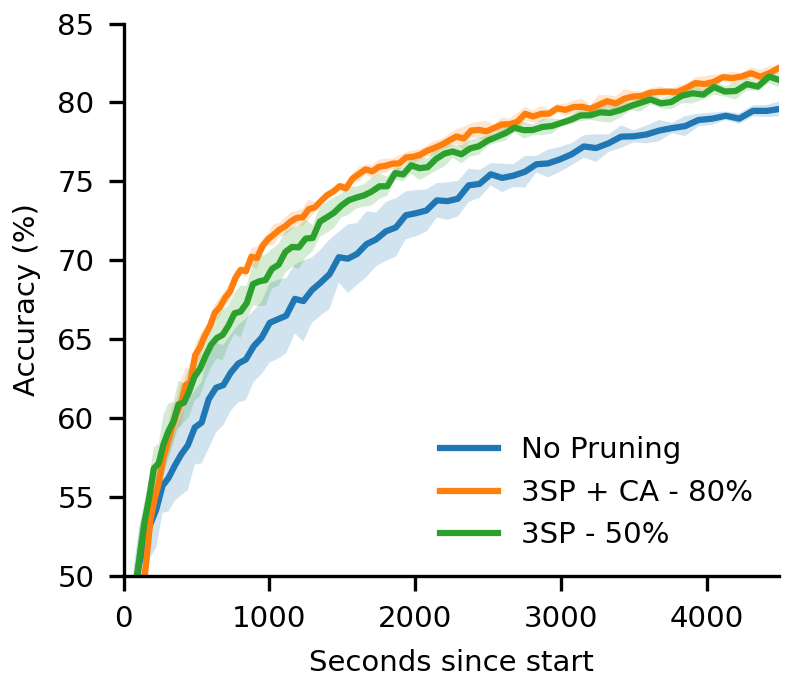}
    \caption{Active Learning CIFAR-10. 3SP model at 50\% and 3SP-CA at 80\% FLOPs reduction have better accuracy than a full model within a time-budget because they train, and therefore acquire data, more quickly.}
    \vspace{-4mm}
    \label{fig:al}
\end{wrapfigure}

\subsection{Active Learning with 3SP Pruning}\label{sec:active_learning}
In some settings, we have a large pool of unlabelled data for which we can request labels, but the labelling process is costly (for example, requiring highly-trained experts).
Active learning \citep{Settles2010} attempts to pick the most informative datapoints out of the pool to reduce the labelling cost.
Unfortunately, most active learning approaches require retraining the model on the labelled set before acquiring new points.
This can mean that the model must be trained tens or hundreds of times during the acquisition process.
We show that a 3SP model can reach a higher accuracy within a time-budget than an unpruned model (Figure \ref{fig:al}).
Even though pruning hurts model performance, the fact that we prune before training in a structured way lets us train much more quickly.
This lets the 3SP model acquire more data within the same time-budget, which allows higher accuracy.
Full experimental details are provided in Appendix \ref{app:active_learning}.

\begin{table}[t]
    \centering
    \caption{Effect of the computational cost reduction on training/inference time.
        Reducing FLOPs directly reduces training time and prediction time, and indirectly reduces the model size.
        These results are obtained using the VGG-19 model, CIFAR10 dataset, and a GTX 1080 Ti GPU.
        The prediction time is measured using a batch with a single element on a i7 8700K CPU.
        The pruning time is equal to one step in the original model.}
    \begin{tabular}{lccc}
        \toprule
                                      & Original          & 50\% FLOPs Reduction        & 80\% FLOPs Reduction          \\
        \midrule
        Training Time per Epoch (GPU) & 15 s              & 8 s (\textasciitilde2x)     & 4 s (\textasciitilde4x)       \\
        Prediction Time (CPU)         & 12.8 ms           & 4.2 ms  (\textasciitilde3x) & 2.6 ms  (\textasciitilde5x)   \\
        Model Size                    & 87 MB             & 12 MB  (\textasciitilde7x)  & 3.3 MB   (\textasciitilde26x) \\
        Pruning Time                  & --                & 41 ms                       & 41 ms                         \\
        Accuracy (CIFAR-10)           & 93.6\% $\pm$ 0.04 & 93.1\% $\pm$ 0.09           & 90.9 $\pm$ 0.09\%             \\
        \bottomrule
    \end{tabular}
    \label{table:compute_comparison}
    \vspace{-2mm}
\end{table}

%!TEX root = ../neurips_2020.tex

\section{Alternative Pruning Criterion - GraSP}
\label{sec:grasp}

\citet{wang_picking_2019} propose to keep network parameters that contribute the most to loss reduction during optimization.
Instead of using the change in loss as a pruning score, like SNIP, they use the predicted change in the magnitude of the gradient with respect to the weights, approximated with a first-order Taylor expansion.
In order to avoid finding a mask that satisfies their objective trivially by creating a very large loss, they strongly smooth outputs so that removing weights cannot greatly change the loss.
Although \citet{wang_picking_2019} motivate their method by the fact that their score measures the interaction between weights, we observe that it is still a first-order method and does not assess the interaction between the decision to \emph{prune} multiple weights.

The naive form of GraSP cannot be directly applied to structured models because GraSP computes gradients with respect to weights directly, not mask variables.
But we consider a structured method inspired by GraSP which instead computes gradients with respect to mask tensor entries.
We found in our experiments that this objective performed substantially worse than one based on SNIP.
On CIFAR10, structured GraSP achieves accuracies of 91.96\%, 91.6\%, and 90.92\% at 80-90-95\% prune ratios respectively, worse than Uniform pruning (compare to Table~\ref{table:CIFAR10}).

\begin{figure}[t]
    \centering
    \begin{subfigure}[b]{0.49\textwidth}
        \centering
        \includegraphics[width=\textwidth]{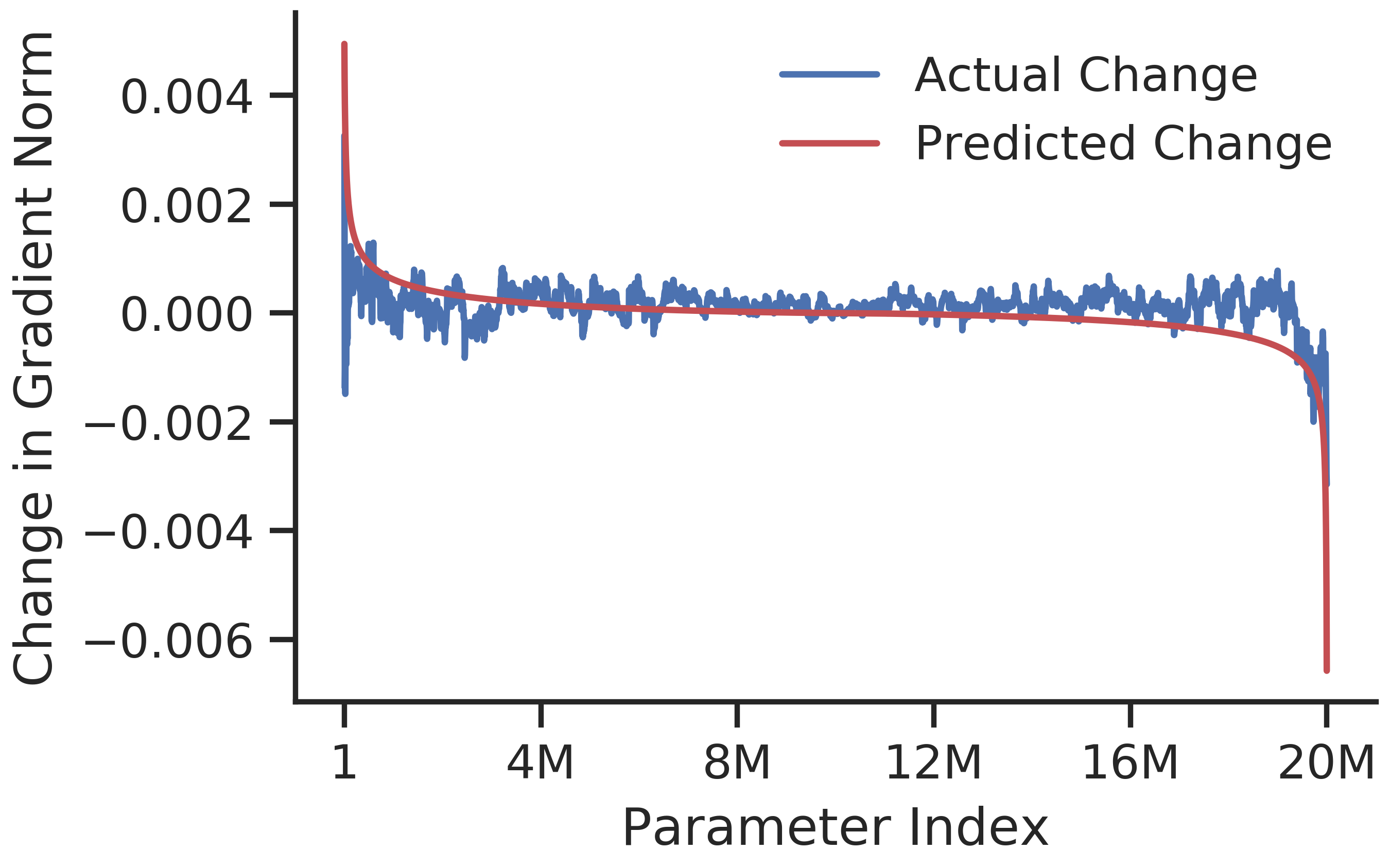}
        \caption{Unstructured}
    \end{subfigure}
    \hfill
    \begin{subfigure}[b]{0.49\textwidth}
        \centering
        \includegraphics[width=\textwidth]{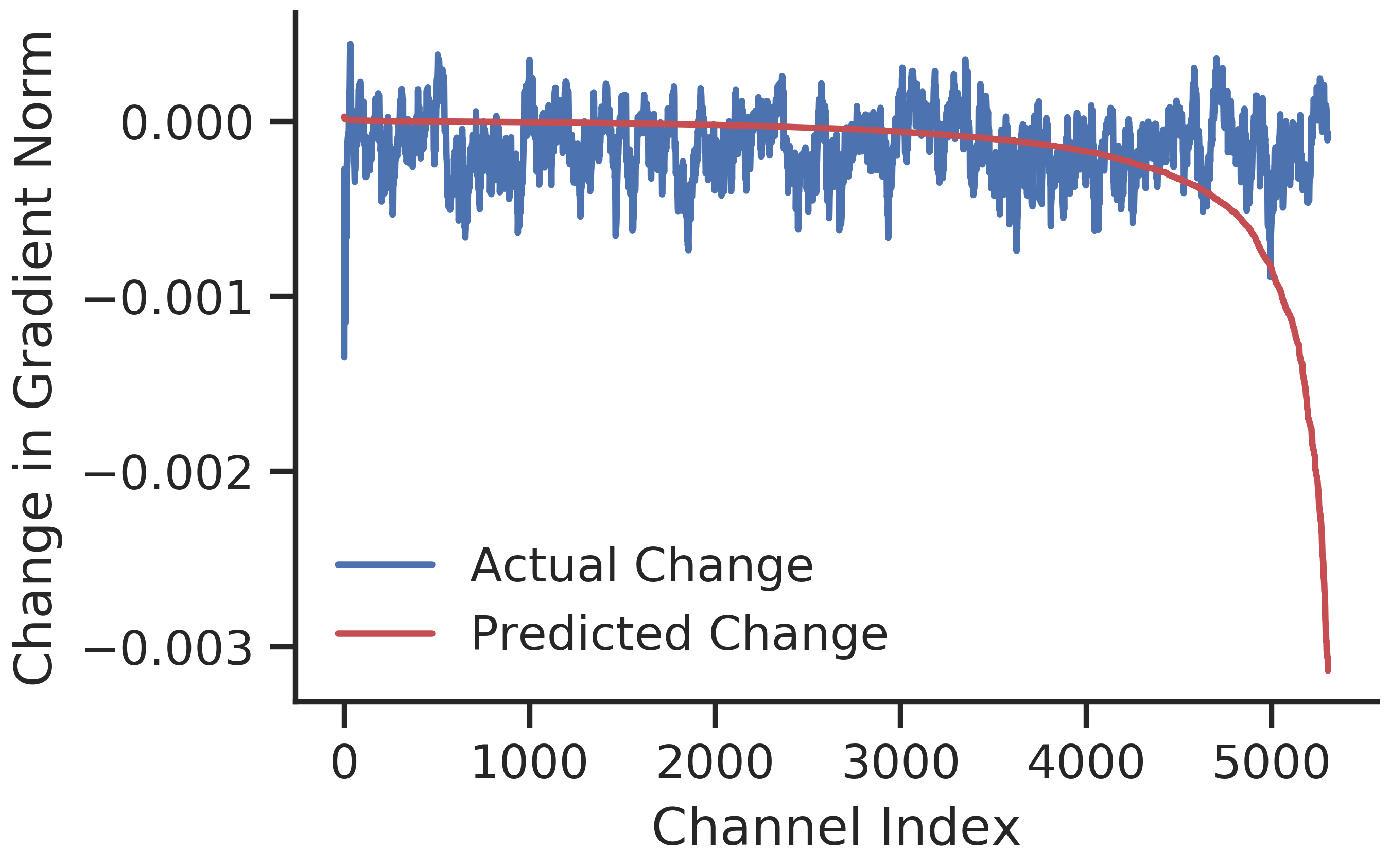}
        \caption{Structured}
    \end{subfigure}
    \caption{
        Evaluating approximation accuracy of GraSP objective for structured and unstructured pruning of VGG-19.
        We compare the predicted change in the gradient-norm given by the Taylor expansion (scores) with the actual change in the gradient-norm (normalized).
        The unstructured predictions have signal, but in the structured setting they are too noisy.
    }
    \label{fig:grasp}
\end{figure}

We believe that this is because estimates of the GraSP-style objective are too noisy in the structured setting, and so assumption 2 considered in \S\ref{sec:method} is violated.
Similar to the experiment in Figure~\ref{fig:pred_vs_actual_snip}, we compare the predictions implied by the gradient to the actual effect on the objective.
Figure~\ref{fig:grasp} shows that in the unstructured case, there is a correlation between the approximated change in the objective and the actual change.
However, the structured prediction is uncorrelated with the actual change.

%!TEX root = ../neurips_2020.tex

\section{Related Work}\label{sec:related_work}

Large neural networks are overparameterized, resulting in extensive efforts to reduce model size through distillation \citep{hinton2015distilling}, pruning \citep{reed_pruning_1993} or quantization \citep{gong_compressing_2014, rastegari2016xnor}.
All of these methods aim to make networks cheaper to evaluate by reducing compute cost, memory usage, or enabling hardware-efficient implementations.
However, none of these works focusses on reducing computation cost both during and after training.
SNIP \citep{lee2018snip} and GraSP \citep{wang_picking_2019} are the first to look at reducing the number of parameters before training but are unable to substantially reduce compute cost due to being unstructured methods.
We discuss these methods extensively in \S\ref{sec:background}.

\citet{pruning_from_scratch_wang} similarly proposes a channel-wise mask as in 3SP, and directly optimizes it using a sparsity penalty similar to \citet{liu2017learning}.
The proposed algorithm requires 10 epochs of training with a computational cost similar to the original model in order to do pruning on CIFAR-10, whereas we compute pruning scores with just one batch.
We consider this a prohibitive additional cost and do not explicitly compare.

Within the pruning after training literature, there are a number of methods that explicitly consider computational cost and others that apply structured pruning.
\citet{gordon_morphnet_2018} explicitly regularize FLOPS while training with a sparsity penalty, while \citet{veniat_learning_2018} and \citet{theis_faster_2018} adopt a similar method motivated as approximate constrained optimization.
\citet{he_amc_2019} frame model crafting as a reinforcement learning problem where the reward is based on compute usage.
Many methods for pruning after training impose a structured pruning mask, which results in compute savings \citep{li_pruning_2016, he_channel_2017, liu2017learning, thinet}.
In particular \citet{eigendamage} obtains very strong results but needs 160 epochs of training a full-sized network and another 160 epochs of fine-tuning.

Neural Architecture Search (NAS) is related to our work insofar as it optimizes a network architecture before training.
However, our work requires only one forwards-backwards pass on a single batch to prune, rather than requiring extensive retraining.
In general, NAS methods have been accused of consuming enormous amounts of energy \citep{strubell_energy_2019}, counter to the goal of this paper.

%!TEX root = ../neurips_2020.tex

\section{Conclusion}\label{sec:conclusion}

We introduced 3SP; a single shot structured pruning technique that is applied before training.
Our results show that it is possible to speed up training by 2x, and inference by 3x while losing only 0.5\% accuracy.
Aside from the reduced time and energy needed to obtain a converged model, this is important in settings where training speed is a bottleneck on performance, such as active learning.
We have further shown that the approximations in SNIP and our structured extension are reasonable and that making 3SP compute aware leads to even more speed ups, at the expense of accuracy.
In general, our work has shown for the first time, that structured pruning before training is feasible and can be an exciting way to reduce wasteful training cost and practioner architecture selection time.

\section*{Acknowledgements}
We thank Aidan Gomez, Bas Veeling and Yee Whye Teh for helpful discussions and feedback.
We also thank others in OATML and OxCSML for feedback at several stages in the project.
JvA/MA are  grateful  for  funding  by  the  EPSRC (grant  reference EP/N509711/1 and EP/R512333/1, respectively).
JvA is also grateful for funding by Google-DeepMind, while MA is supported by Arm Inc.
SF gratefully acknowledges the Engineering and Physical Sciences Research Council for their grant administered by the Centre for Doctoral Training in Cyber Security at the University of Oxford.

%!TEX root = ../neurips_2020.tex

\section*{Broader Impact}

Our work's key impact is the development of new compute-aware pruning techniques.
We think this is broadly beneficial because of the reduction in energy consumption and carbon dioxide emissions it makes possible.
We can separate the impacts into immediate beneficial applications, medium-term extensions, and possible unintended consequences.

\paragraph{Immediate benefits.}
Modern machine learning methods consume huge amounts of energy.
\citet{strubell_energy_2019} have noted that this is not just a problem at production scale, large amounts of energy are spent on training models and especially on architecture search.
We provide a mechanism for creating compute-efficient models before training, which reduces energy consumption in training models.
Moreover, unlike neural architecture search methods, we require just one batch of backpropagation on the target dataset before training, rather than huge numbers of model evaluations.
The immediate impact of our paper is therefore to help researchers and labs reduce their energy consumption and carbon dioxide emissions.

In addition, because of how expensive neural architecture search is, it is mostly only available to large companies.
Our method will be especially useful in resource-poor contexts where teams do not have large datacenters and need to be as efficient as possible, for example in some less economically developed countries.

\paragraph{Medium-term extensions.}
Beyond our direct method, we introduce the framing of optimizing pruning in compute-space.
We hope very much that future work can adopt this mindset and create further reductions in energy consumption.
It may also be useful for some groups to consider pruning in energy-space, however this is too hardware-dependent to be appropriate for a paper of our generality.

\paragraph{Unintended Consequences}
We do not believe there are adverse distributional consequences of our work, in general it should make it more possible for less economically developed actors to achieve results closer to those that only the largest companies can currently achieve.

However, it is possible that improving the computational efficiency of deep learning will increase overall demand for deep learning, thereby increasing overall energy consumption (an instance of the Jevons Paradox \citep{jevons_coal_1865}).
We hope that in such a situation, the benefits to humanity from the increased use of deep learning are worth the greater overall cost.

\newpage
\bibliographystyle{plainnat}
\bibliography{references}

\newpage
\appendix
%!TEX root = ../neurips_2020.tex

\section{Experimental setup}\label{app:architecture}
For VGG-19 we use an architecture of five blocks, with 2x2 max-pooling layers in between the blocks.
The first two blocks consist of two conv-BN-relu operations, with 64 and 128 channels, respectively.
The last three blocks consist of three conv-BN-relu operations and are of width 256, 512 and 512.
After our model we use average pooling to reduce spatial dimensions to 2 by 2.
This is followed by three linear layers with 1024, 512, and number of classes as number of hidden units.
We use relu activation functions throughout the model.
During structured pruning, we consider all elements of the model prunable except the output of the final linear layer and the input of the first convolution layer.
We train our model for a fixed 160 epochs, with an initial learning rate of 0.1, which is halved at epoch 80 and again at 120.
We preprocess CIFAR-10 by doing mean and standard deviation normalization, random horizontal flips, and random 4-pixel pads followed by a 32x32 crop.
We report results on the CIFAR-10 test set.
This is the same experimental setup as used in \citet{wang_picking_2019}; we did not consider other hyperparameters.
We use the default initialization method of Pytorch 1.4 for the weights, which is based on \citet{he2015delving}.
All results are based on 5 training/evaluation runs with different random seeds.

For all full model measurements of computational cost in this paper, we utilize the \texttt{pytorch-OpCounter}\footnote{Available from \url{https://github.com/Lyken17/pytorch-OpCounter}} package, which takes into account all operations of the model, excluding preprocessing.
\section{Wall Clock Time and Model Size Experiments}\label{app:wall_clock}
We performed the experiments in Table \ref{table:compute_comparison} using the VGG-19 described in Appendix \ref{app:architecture}.
The training time per epoch was measured as the average of 5 epochs, excluding the first epoch.
The prediction time is the average prediction time per data point after going through an epoch of data one by one.
The model size is determined by saving the parameters to disk.
Pruning time is determined by doing a forward pass with a single batch of data in the original model.
The time necessary for adjusting the architecture is excluded but would add some overhead (under 10 ms) in practice.
The 50\% FLOPs reduction model is 3SP without compute aware.
The 80\% FLOPs reduction model is 3SP with compute aware.
\begin{wrapfigure}{r}{0.48\textwidth}
    \centering
    \vspace{-2mm}
    \captionsetup{labelfont={bf}}
    \includegraphics[width=2.64in,height=2.3in]{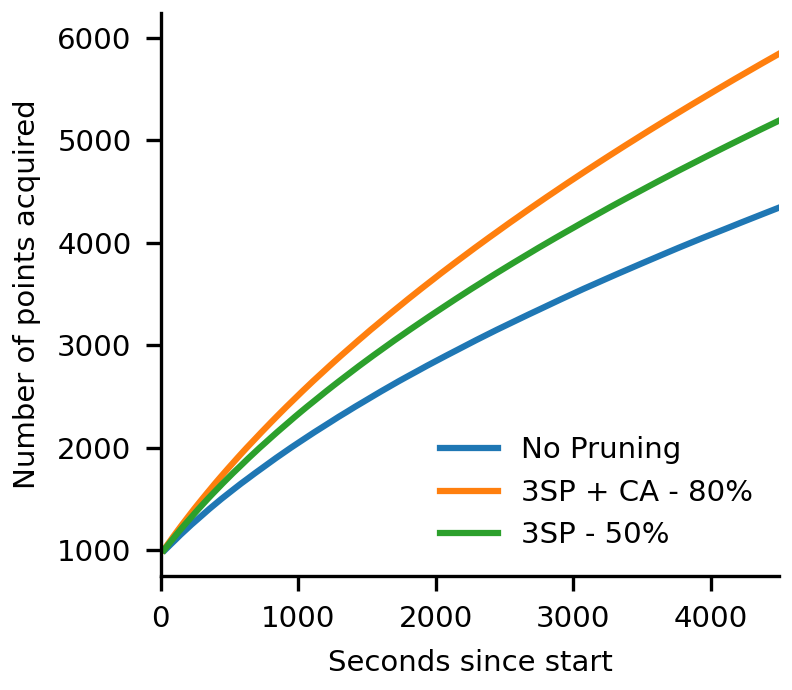}
    \caption{Active Learning CIFAR-10. 3SP model at 50\% and 3SP-CA at 80\% FLOPs reduction get to acquire more data within a time-budget because they train faster.}
    \vspace{-15mm}
    \label{fig:al_data}
\end{wrapfigure}
\section{Active Learning Experiments}\label{app:active_learning}
In active learning \citep{Settles2010}, we want to be efficient about manually labeling data and only acquire labels for the most informative data points.
It is also referred to as a `human in the loop' approach to data labeling.
One begins with a small training dataset, and obtains labels only for the most informative datapoints in the unlabeled set.
In our case, we start with 100 examples per class, leading to a start dataset of size 1000 for CIFAR-10.
We train the model on the current labelled dataset and subsequently use it to estimate which datapoints in the original CIFAR-10 training set (the `unlabeled set') would be most informative to acquire a label for.
We use a common proxy for informativeness: the entropy of the softmax distribution $H(y) = \sum_i p(y_i|\mathbf{x}) \log p(y_i|\mathbf{x})$, where $p(y|\mathbf{x})$ represents the output distribution for a particular data point $\mathbf{x}$ \citep{gal_deep_2017}.
We select 50 datapoints from the unlabeled pool in each acquisition step, and add these (including labels) to the training set and restart training (using the final state of the model before acquiring more data).
We repeat this process until the time runs out.
We obtain data from the CIFAR-10 training set and report results on the CIFAR-10 test set.
We prune the models once at the beginning of the active learning process.
We start the timer after we acquired the 1000 initial points and are done with pruning.
The experiments were done using the VGG-19 architecture and training settings described in Appendix \ref{app:architecture}.
For 3SP - 50\%, we pruned 50\% of the FLOPS of the model, without compute awareness.
For 3SP + CA - 80\%, we pruned 80\% of the FLOPS, but with compute awareness.
We repeat the experiments 5 times, and show one standard deviation errors.

Throughout active learning, the pruned models train more quickly.
This means that they are able to select points for inclusion more quickly (see Figure \ref{fig:al_data}).
3SP + CA sees almost 50\% more data than the un-pruned model within the time-budget.

\end{document}